\documentclass{article}
\usepackage{arxiv}
\usepackage[utf8]{inputenc} 
\usepackage[T1]{fontenc}    
\usepackage{hyperref}       
\usepackage{url}            
\usepackage{booktabs}       
\usepackage{amsfonts}       
\usepackage{nicefrac}       
\usepackage{microtype}      
\usepackage{lipsum}
\usepackage{graphicx}
\usepackage{amsmath}
\usepackage{multirow}
\graphicspath{ {./images/} }

\title{Integrating Reinforcement Learning and AI Agents for Adaptive Robotic Interaction and Assistance in Dementia Care}

\author{
 Fengpei Yuan \\
  Robotics Engineering\\
  Worcester Polytechnic Institute\\
  Worcester, MA 01609 \\
  \texttt{fyuan3@wpi.edu} \\
   \And
 Nehal Hasnaeen \\
  University of Tennessee\\
  Knoxville, TN 37996 \\
  \texttt{shasnaee@vols.utk.edu} \\
  \And
 Ran Zhang \\
  University of North Carolina\\
  Charlotte, NC 28223 \\
  \texttt{rzhang8@charlotte.edu} \\
  \And
  Bryce Bible \\
  University of Tennessee\\
  Knoxville, TN 37996 \\
  \texttt{bbible3@vols.utk.edu} \\
  \And
  Joseph Riley Taylor \\
  University of Tennessee \\
  Knoxville, TN 37996 \\
  \texttt{jtayl219@vols.utk.edu} \\
  \And
  Hairong Qi \\
  University of Tennessee\\
  Knoxville, TN 37996 \\
  \texttt{hqi@utk.edu} \\
  \And
  Fenghui Yao \\
  Tennessee State University\\
  Nashville, TN 37209 \\
  \texttt{fyao@Tnstate.edu}
  \And
  Xiaopeng Zhao \\
  University of Tennessee\\
  Knoxville, TN 37996 \\
  \texttt{xzhao9@utk.edu}
}

\begin{document}
\maketitle

\begin{abstract}
  This study explores a novel approach to advancing dementia care by integrating socially assistive robotics, reinforcement learning (RL), large language models (LLMs), and clinical domain expertise within a simulated environment. This integration addresses the critical challenge of limited experimental data in socially assistive robotics for dementia care, providing a dynamic simulation environment that realistically models interactions between persons living with dementia (PLWDs) and robotic caregivers. The proposed framework introduces a probabilistic model to represent the cognitive and emotional states of PLWDs, combined with an LLM-based behavior simulation to emulate their responses. We further develop and train an adaptive RL system enabling humanoid robots, such as Pepper, to deliver context-aware and personalized interactions and assistance based on PLWDs' cognitive and emotional states. The framework also generalizes to computer-based agents, highlighting its versatility. Results demonstrate that the RL system, enhanced by LLMs, effectively interprets and responds to the complex needs of PLWDs, providing tailored caregiving strategies. This research contributes to human-computer and human-robot interaction by offering a customizable AI-driven caregiving platform, advancing understanding of dementia-related challenges, and fostering collaborative innovation in assistive technologies. The proposed approach has the potential to enhance the independence and quality of life for PLWDs while alleviating caregiver burden, underscoring the transformative role of interaction-focused AI systems in dementia care.
\end{abstract}

\section{Introduction}
Alzheimer’s disease (AD) and related dementias (ADRD) account for $60\text{--}80\%$ of dementia cases, affecting approximately 6.9 million Americans aged 65 and older, with projections rising to 13.9 million by 2060 \cite{AA2024diseasefact}. Persons living with AD/ADRD (PLWDs) often face challenges such as short-term memory loss, attention deficits, and negative emotions (e.g., anxiety, anger, and frustration), which disrupt their ability to initiate or perform daily routine tasks properly \cite{AA2024diseasefact,woods2001discovering}. As a result, PLWDs require extensive care, companionship, and assistance with activities of daily living (ADLs) like meal preparation, shopping, transportation, and money management \cite{langbaum2023recommendations}. However, caregiving for dementia is demanding, with caregivers typically providing over 21 hours of weekly care for four years or more, leading to increased depression, stress, and physical health decline \cite{AA2024diseasefact}. This growing need for dementia care exacerbates an already critical shortage of skilled caregivers \cite{olivari2020public}.

To address this challenge, AI-powered technologies, including computers, smartphones, and robotics, have emerged as promising solutions for enhancing the qualify of life of both PLWDs and their caregivers through monitoring, companionship, and assistance~\cite{moyle2019promise,Yuan2021FRA,woods_social_2021-1,yuan_cognitive_2023,pappada2021assistive}.
Among these technologies, socially assistive robotics (SAR) has shown particular promise in dementia care due to two key advantages: (1) Physical embodiment, which has been shown to enhance task initiation and engagement for PLWDs \cite{lee2006physically,ghafurian2021social}, and 2) Multimodal interaction capabilities (e.g., voice, gestures, body movement, and graphical interfaces) that enhance task engagement and comprehension for PLWDs in social activities and therapies \cite{ghafurian2021social,Yuan2021FRA}. For instance, robotic pets like PARO have demonstrated anxiety-reducing effects and promoted social engagement despite ethical concerns and high costs \cite{hung_benefits_2019-1}. Humanoid robots such as Pepper, Pearl, Mario, and TIAGo Iron offer cognitive stimulation, reminders, and personalized health monitoring to support independence and well-being \cite{tombot2024, cocsar2020enrichme, yuan_cognitive_2023, yuan_simulated_2022, woods_social_2021-1, portugal_socialrobot_2015, wu_socially_2021,yuan2024social}. Studies also emphasize the importance of safety features, voice communication, and personalization in AI-assisted dementia care \cite{yuan2022assessing}.

Despite these advancements, SAR applications for assisting with ADLs remain limited. Most existing studies rely on tele-operated robots, requiring human intervention, which restricts adaptability, autonomy, and scalability \cite{ghafurian2021social,yuan2024social}. For instance, Wang et al.~\cite{wang2017robots} used a tele-operated robot to provide stepwise prompting for tasks like hand-washing and coffee preparation, highlighting the potential of SAR to support PLWDs and caregivers. However, tele-operation approaches fail to fully leverage robotic autonomy, limiting their adaptability to diverse ADL scenarios or unique cognitive and affective states of PLWDs \cite{woods2001discovering}. A significant barrier to transitioning from tele-operated to autonomous robotic systems is the limited availability of datasets reflecting the nuanced behaviors and emotional states of PLWDs during daily activities. This scarcity hinders the development of context-aware, emotion-aware, autonomous decision-making in robotic systems. These challenges highlight the need for novel human-computer interaction approaches that can bridge the gap between current tele-operated systems and fully autonomous robotic assistance, while maintaining the critical aspects of social engagement and personalization that make SAR effective for dementia care.

To address these gaps, we introduce an innovative framework that integrates reinforcement learning (RL) and large language models (LLMs)—such as ChatGPT (GPT-4o)—to develop AI agents for SAR and human-robot interaction in dementia care, leveraging the extensive background knowledge about AD/ADRD and dementia care embedded in these LLMs. Our main contributions include: (1). an open-source simulator to model realistic PLWD-caregiver interactions during ADLs, generating synthetic behavioral data that reflect the nuanced cognitive and affective states of PLWDs. This open-source simulator provides a valuable tool for RL researchers to advance robotic decision-making and assistive strategies, and (2) RL-based strategies developed for autonomous, adaptive, person-centered, practical interventions for robots and other assistive agents (e.g., smartphones, computers) to enhance PLWDs’ engagement and autonomy in daily activities. The proposed framework fosters collaborative development and iterative improvement, supporting the broader human-computer interaction (HCI) and AI research community.

The remainder of this paper is structured as follows. Section 2 details the development of the AI agent \textit{PLWD}, incorporating LLMs, prompt engineering, and clinical insights. Section 3 describes the RL-based agent \textit{Robot Caregiver}. Section 4 presents experimental results, including action selection via RL and the simulated interactions between \textit{PLWD} and \textit{Robot Caregiver}. Ethical considerations and future improvements are also discussed. Section 5 concludes the paper.

\section{Agent PLWD}
\begin{figure}[h]
    \centering
    \includegraphics[width=0.77\textwidth]{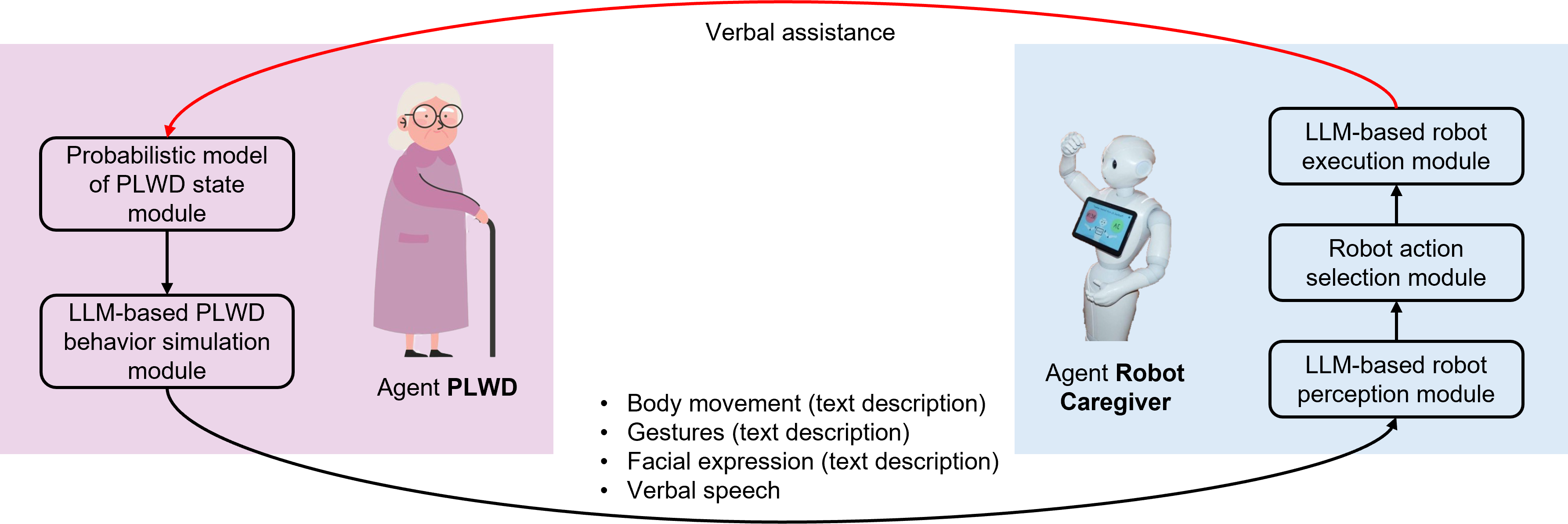}
    \caption{Illustration of the system architecture for our two AI agents, \textit{PLWD} and \textit{Robot Caregiver}. The agent \textit{PLWD} is built with two modules. The agent \textit{Robot Caregiver} consists of three modules for perception, decision-making, and action execution, correspondingly.}
    \label{fig:systemArchitecture_TwoAIAgents}
\end{figure}

Inspired by the potential of robot caregiver for individuals with dementia at their home and the lack of real-world dataset in this application, we build a simulated environment including two AI agents, \textit{PLWD} and \textit{Robot Caregiver}. 
Fig.~\ref{fig:systemArchitecture_TwoAIAgents} illustrates the system architecture of the two simulated AI agents and their dynamic interactions.
In the simulation, the agent \textit{PLWD} performs daily routine tasks and the assistive agent \textit{Robot Caregiver} observes PLWD's status and provides assistance accordingly, to help PLWD complete the task.
The agent \textit{PLWD} consists of two modules, an abstract-level statistical model for PLWD's cognitive and emotional states and an LLM-based PLWD behavior simulation model which models the realistic behavior, both verbal and nonverbal, during performing the task.
The agent \textit{Robot Caregiver} includes three modules, robotic perception, decision-making, and action execution. The modules of perception and action execution are leveraging the powerful LLM model. More details on the development of agent \textit{Robot Caregiver} are given in the next section. 

\begin{figure}[h]
    \centering
    \includegraphics[width=0.47\textwidth]{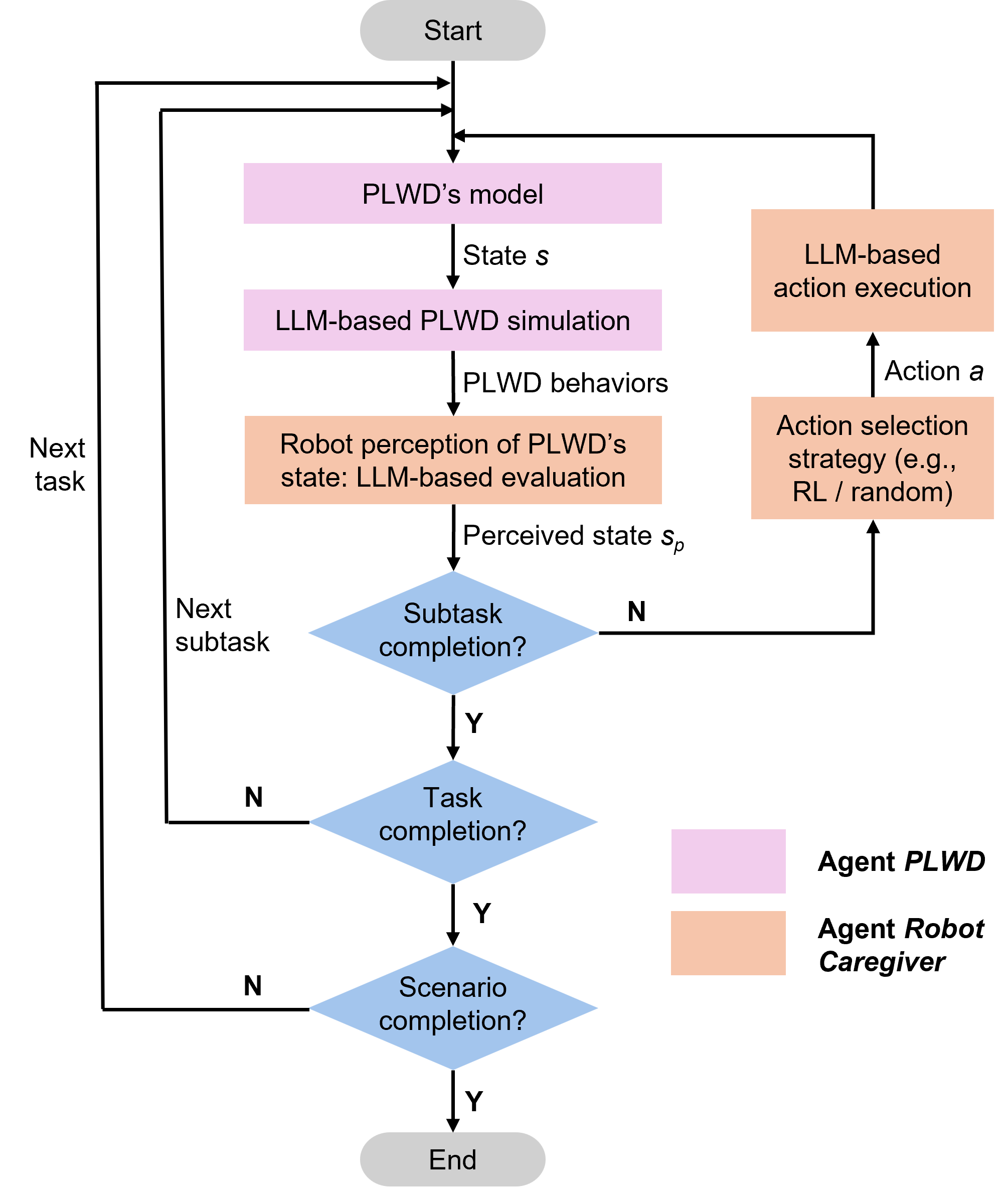}
    \caption{The flowchart illustrates the dynamic interaction between the two AI agents, a person living with dementia (PLWD) and the robot caregiver, during a daily activity scenario. The pink and orange blocks represent the simulation of PLWD and the robot caregiver, correspondingly.}
    \label{fig:framework}
\end{figure}

The dynamic interaction between the two agents are shown in Fig.~\ref{fig:framework}. The process begins with statistical modeling, which uses probabilistic methods to simulate PLWD's cognitive and affective states, informed by clinical observations and theoretical insights. The LLM-based behavior simulation module converts the numerical outputs into comprehensive textual descriptions, providing nuanced depictions of the PLWD's cognitive and emotional states for the robot’s perception system.
Our current simulator will focus on the advanced ADLs with a cognitive emphasis, for example, shopping. For the convenience of discussion, we are introducing concepts of scenarios, tasks, subtasks, trials, and timesteps in the simulator. Their definitions and examples are listed in Table~\ref{tab:termDefinitions}.

\begin{table}
\renewcommand{\arraystretch}{1.5}
\caption{Terms and Definitions used in our Simulating Context\label{tab:termDefinitions}}
\centering
\begin{tabular}{p{1.2cm}p{5.8cm}p{6.8cm}}
\toprule
{\textbf{Term}}& \hfil{\textbf{Definition}} & \hfil{\textbf{Example}} \\
\midrule
\textit{Scenario} & The broad context or setting that includes various tasks, starting from an initial state and ending at a terminal state & Shopping scenario.\\
\textit{Task} & Specific objectives within a scenario, often complex and divisible into subtasks & \textit{Task 1}: Select 3 items on the shopping list correctly; \textit{Task 2}: Select the correct cash for the 3 grocery items.\\
\textit{Subtask} & Smaller, manageable components of a task & In Task 1, there are \textit{Subtask 1}: identify item one; \textit{Subtask 2}: pick up item one and gather it in a location; \textit{Subtask 3}: identify item two, etc.\\
\textit{Trial} & A single attempt or iteration where PLWD tries to complete the specified subtask, with or without external assistance & In a subtask of Task 1, if PLWD picks one item and pauses, a caregiver might remind them by asking, "Is there anything missing?" prompting PLWD to check the shopping list.\\
\textit{Timestep} & Time steps of interaction between PLWD and a caregiver, occurring under the scenario & The total timesteps of a scenario equal the sum of trials over all subtasks and tasks.\\
\bottomrule
\end{tabular}
\end{table}

\subsection{Probabilistic Model of PLWD States}
We have developed a statistical model to simulate the probability of performance breakdown in PLWDs during performing ADLs, with and without external assistance. The performance breakdown is particularly manifested as PLWD's cognitive and affective statuses, including forgetfulness, confusion, anger, and disengagement, which are affected by ADRD such as cognitive deficits. All these four statuses might hinder PLWDs' success, safety, independence, and adequacy in performing the daily tasks. In other words, all these moments could be moments when the socially assistive robot assists PLWDs in task completion. In this work, we will mainly focus on persons with early- or moderate-stage dementia, the population who typically shows performance breakdown from a cognitive perspective.

The potential external assistance that might be provided to PLWD is designed by adopting the hierarchy of assistive levels for PLWDs from the Performance of Self-care Skills (PASS) \cite{chisholm2014evaluating}. PASS has been widely used by occupational therapists to evaluate individuals' capacity to perform ADLs and live independently and safely in the community for people with AD/ADRD \cite{dham2020functional,keleman2021amyloid,yuan2022assessing}. 
In PASS protocol, there are 10 different levels of assistance that could be provided to PLWDs to help them successfully complete their ADLs.
Considering that our current target population includes people with mild or moderate dementia, this paper focuses on assistance from the cognitive and affective perspectives; therefore, the assistance includes `\textit{No Assistance}', `\textit{Verbal Supportive Assistance}' (encouragement),`\textit{Verbal Non-directive Assistance}' (cue to alert), and `\textit{Verbal Directive Assistance}' (instruction). Their definitions and examples are given in Table \ref{tab:assistance_type}. 

\begin{table}
\renewcommand{\arraystretch}{1.5}
\caption{Definitions and Examples of Assistance Types\label{tab:assistance_type}}
\centering
\begin{tabular}{p{2.8cm}p{6cm}p{4.5cm}}
\toprule
{{Assistance Type}}& {{Definition}} & {{Example}} \\
\midrule
No assistance ($a_0$) & No assistance is provided from the robot & \\
Verbal supportive \newline assistance ($a_1$) & Encouragement to initiate, continue, or complete a task & “{Keep at it”; “Great”}\\
Verbal non-directive assistance ($a_2$) & Cues to facilitate task initiation, continuance, or completion without telling PLWD exactly what to do; often stated in the form of a question & “Is there anything missing?”;
“Can you try another way?”\\
Verbal directive \newline assistance ($a_3$)& Verbal statements to initiate, continue, or complete a task & “Check the recipe again”; “The date needs to be filled in on the check”\\
\bottomrule
\end{tabular}
\end{table}

\begin{figure}[htbp]
\centering
\includegraphics[width=0.4\textwidth]{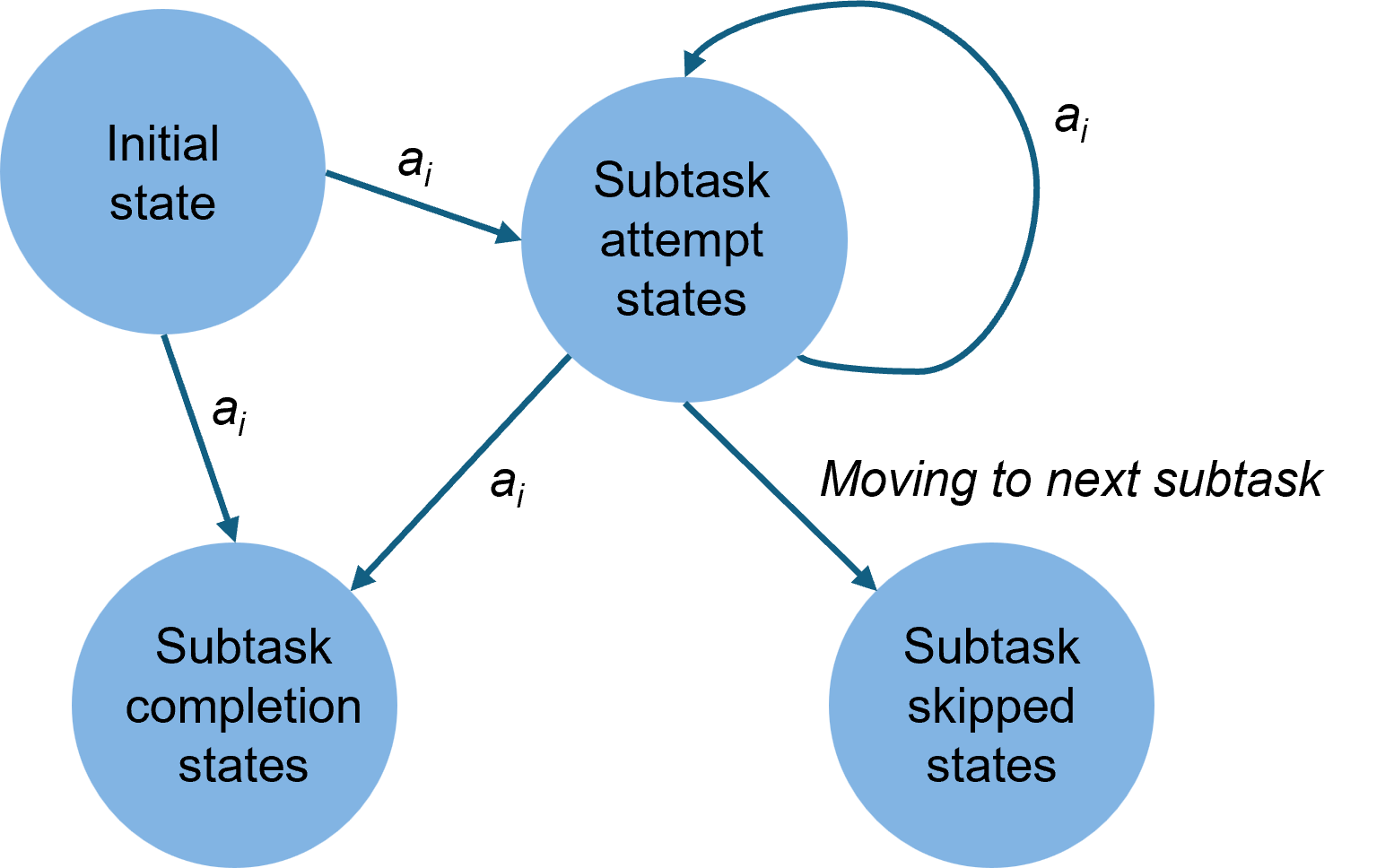}
\caption{A high-level illustration of the state transition model for agent \textit{PLWD}. Action $a_i$ could be any of the four actions, $a_0$-$a_3$.}
\label{fig:stateTransitionDiagram}
\end{figure}

\begin{figure}[htbp]
\centering
\includegraphics[width=0.4\textwidth]{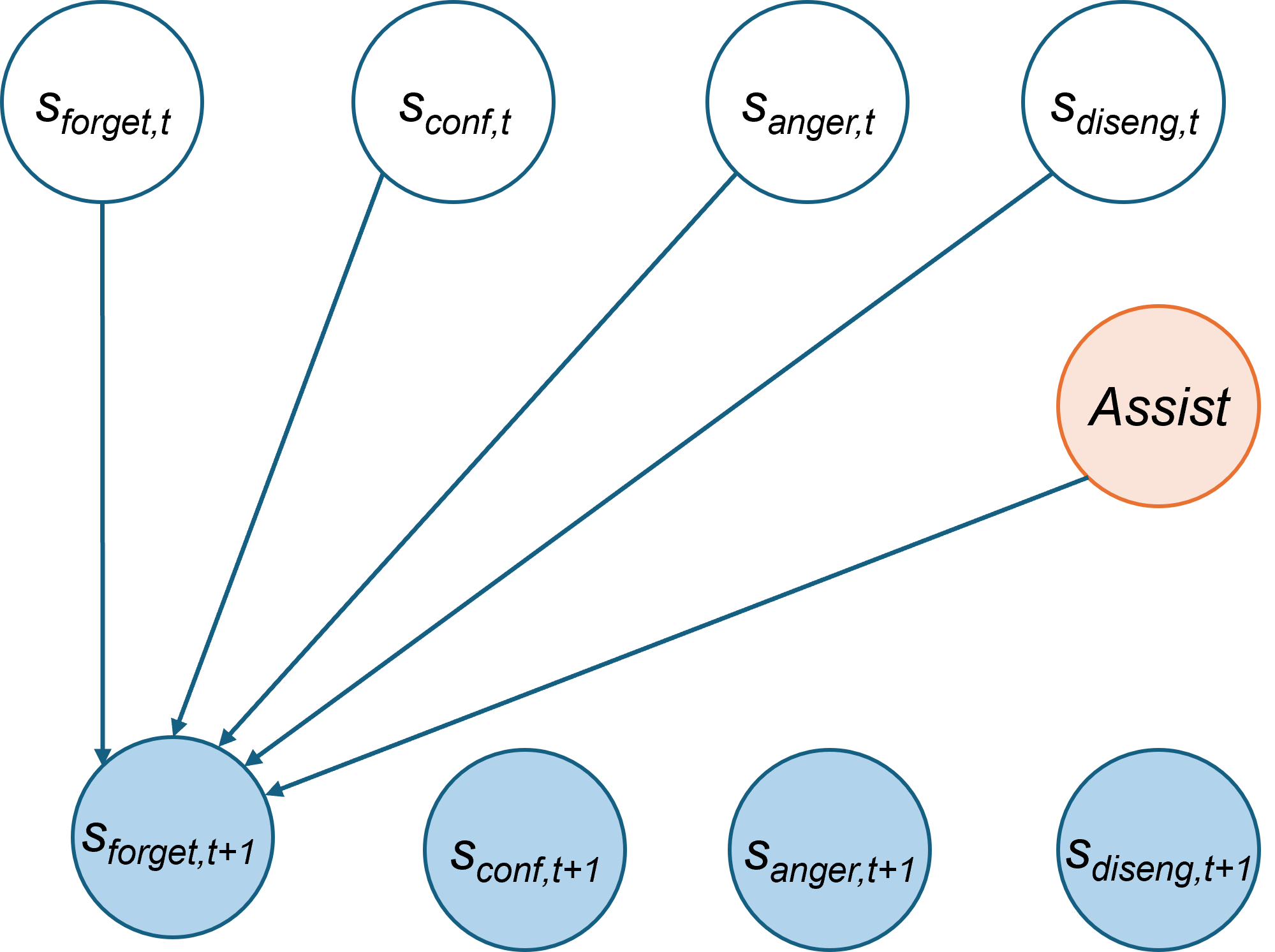}
\caption{A graph illustration of how the dynamics of PLWD's status \textit{Forgetfulness} is simulated. The same mechanics applied to the other three statuses, \textit{Confusion}, \textit{Anger}, and \textit{Disengagement}.}
\label{fig:stateActionInfluenceMechanism}
\end{figure}

Given the terms in Table \ref{tab:termDefinitions} and the four types of assistance in Table \ref{tab:assistance_type}, transitions of PLWD states given the four assistance are illustrated in Fig.~\ref{fig:stateTransitionDiagram}. 
In each subtask, if PLWD successfully finishes the current subtask within $MaxTrial$, with or without assistance from the robot caregiver, PLWD will move to the subtask completion state and next subtask. On the other hand, if PLWD cannot finish the current subtask within $MaxTrial$, PLWD will skip current subtask and move to next subtask. If PLWD skips the current subtask and moves to the next subtask, PLWD's state will transit to a \textit{subtask skipped state}. 
More specifically, the dynamics of PLWD's cognitive and affective statuses when performing the tasks is simulated. The PLWD's dynamic statuses could be influenced by their cognitive deficits, the task and subtask, and external factors (i.e., assistance). For simplification, we use Markov Chain to simulate the dynamics of PLWD's four cognitive and affective statuses, taking into consideration their current cognitive and affective statuses as well as external assistance. Fig.~\ref{fig:stateActionInfluenceMechanism} illustrates the mechanisms of PLWD's dynamic probabilistic transitions. 

Noticeably, our current probabilistic PLWD model only considers general situations, instead of the varieties in dementia population. Here are the four assumptions to simulate PLWD's cognitive and affective statuses. 
The four statuses, \textit{Forgetfulness}, \textit{Confusion}, \textit{Anger}, and \textit{Disengagement}, are modelled as binary (Yes = 1, No = 0).
Each status is assumed to have a base probability of transitioning from No to Yes, and from Yes to No, in the absence of the other three statuses and external assistance. 
At the same time, the dynamics of each status will be influenced by the presence of other three statuses and external assistance (if applicable).
Without external assistance, there is a very small or even 0 probability of transition from Yes to No, which represents a process where dementia patients could become better by themselves. In other words, there is almost a probability of 1 of transition from Yes to Yes, without external assistance. 

\subsubsection{Without external assistance}
Based on the assumptions and qualitative observations of PLWD's cognitive and affective behaviors from previous studies \cite{chisholm2014evaluating,yuan2022assessment}, we simulate the dynamics of PLWD's four statuses using the probabilities in Table~\ref{tab:probabilisticTransition_NoAssistance}. 
While these probabilities are simplifications of the complexity inherent in dementia, they provide a useful framework for understanding the general trends in PLWD's cognitive and emotional state transition, considering factors such as cognition, mood, task nature, and coping mechanisms.

For example, \textit{Forgetfulness} is influenced by \textit{Anger} and \textit{Disengagement}, showing the impact of cognitive-emotional stressors on memory. Recognizing that \textit{Anger} can severely disrupt memory and cognitive functions, it contributes more to other statuses. Our model reflects the strong connection between cognitive impairments (forgetfulness and confusion) and emotional responses (anger), acknowledging that cognitive challenges often lead to emotional distress. Additionally, cognitive and emotional statuses (i.e., \textit{Forgetfulness}, \textit{Confusion}, and \textit{Anger}) can lead to withdrawal or lack of engagement. Confusion, anger, and disengagement exacerbate forgetfulness, with anger having a larger influence due to its intensity, aligning with clinical observations that emotional distress significantly impacts cognitive functions like memory.

\begin{table}
\caption{Probabilistic Transition of PLWD's cognitive and affective status without external assistance\label{tab:probabilisticTransition_NoAssistance}}
\centering
\begin{tabular}{cccccc}
\toprule
\multirow{2}{*}{Status} & \multirow{2}{*}{Base Probability} & \multicolumn{4}{c}{Probability of mutual influence} \\ \cline{3-6}
 &  & $Forgetful_{t+1}$ & $Confused_{t+1}$ & $Angry_{t+1}$ & $Disengaged_{t+1}$ \\
\midrule
$Forgetful_{t}$ & $30\%$ & $99\%$  & $+5\%$  & $+6\%$ & $+2\%$ \\
$Confused_{t}$ & $30\%$ & $+7\%$ &  $99\%$ & $+8\%$ & $+2\%$ \\
$Angry_{t}$ & $5\%$ & $+7\%$ & $+8\%$  & $100\%$ & $+10\%$ \\
$Disengaged_{t}$ & $20\%$ & $+7\%$ & $+10\%$  & $+20\%$ & $99\%$ \\
\bottomrule
\end{tabular}
\end{table}

\subsubsection{With external assistance}
The transition probabilities of the four cognitive and affective statuses will be influenced by the presence of external assistance in Table~\ref{tab:assistance_type}. More specifically,
in the presence of verbal supportive assistance, if PLWD current status \textit{Anger} (\textit{Disengagement}) is Yes, the following status of \textit{Anger} (\textit{Disengagement}) being Yes is $5\%$; Otherwise, it will be 0. 
For the other two statuses, \textit{Forgetfulness} and \textit{Confusion}, their dynamics will follow the rules in {Table~\ref{tab:probabilisticTransition_NoAssistance}}.
In the presence of verbal non-directive assistance or verbal directive assistance, these two types of assistance will not influence the dynamics of emotional states, \textit{Anger} and \textit{Disengagement}, whose dynamics will follow the rules in Table~\ref{tab:probabilisticTransition_NoAssistance}. 
Regarding the statuses \textit{Forgetfulness} and \textit{Confusion}, if their current status is No, the probability of following status being Yes will be $0\%$ and $0\%$, correspondingly.
Comparatively, if their current status is Yes, the probability of following status being Yes will be $40\%$ and $60\%$, correspondingly, given verbal non-directive assistance, and $5\%$ and $5\%$, correspondingly, given verbal directive assistance.

\subsubsection{Subtask skipping}
Acknowledging the complexity of PLWDs' cognitive and affective responses, skipping a task might alleviate some negative emotions but does not guarantee an immediate cognitive and/or emotional reset.
Therefore, the transition probability of \textit{subtask skipped states} follows the following rules: For \textit{Forgetfulness} (\textit{Confusion}), the probability of persistence after skipping a subtask is set at $0.5$ if the PLWD is already forgetful and confused, and $0.2$ if forgetful (confused) but not confused (forgetful), reflecting the impact of emotional stress on memory retention. For the statuses \textit{Anger} and \textit{Disengagement}, a probability of $0.5$ has been assigned for their persistence after subtask skipping, recognizing the potential for these emotions to either continue or abate. 

\subsection{LLM-based PLWD Behavior Simulation}
\begin{figure}[htbp]
\centering
\includegraphics[width=0.9\textwidth]{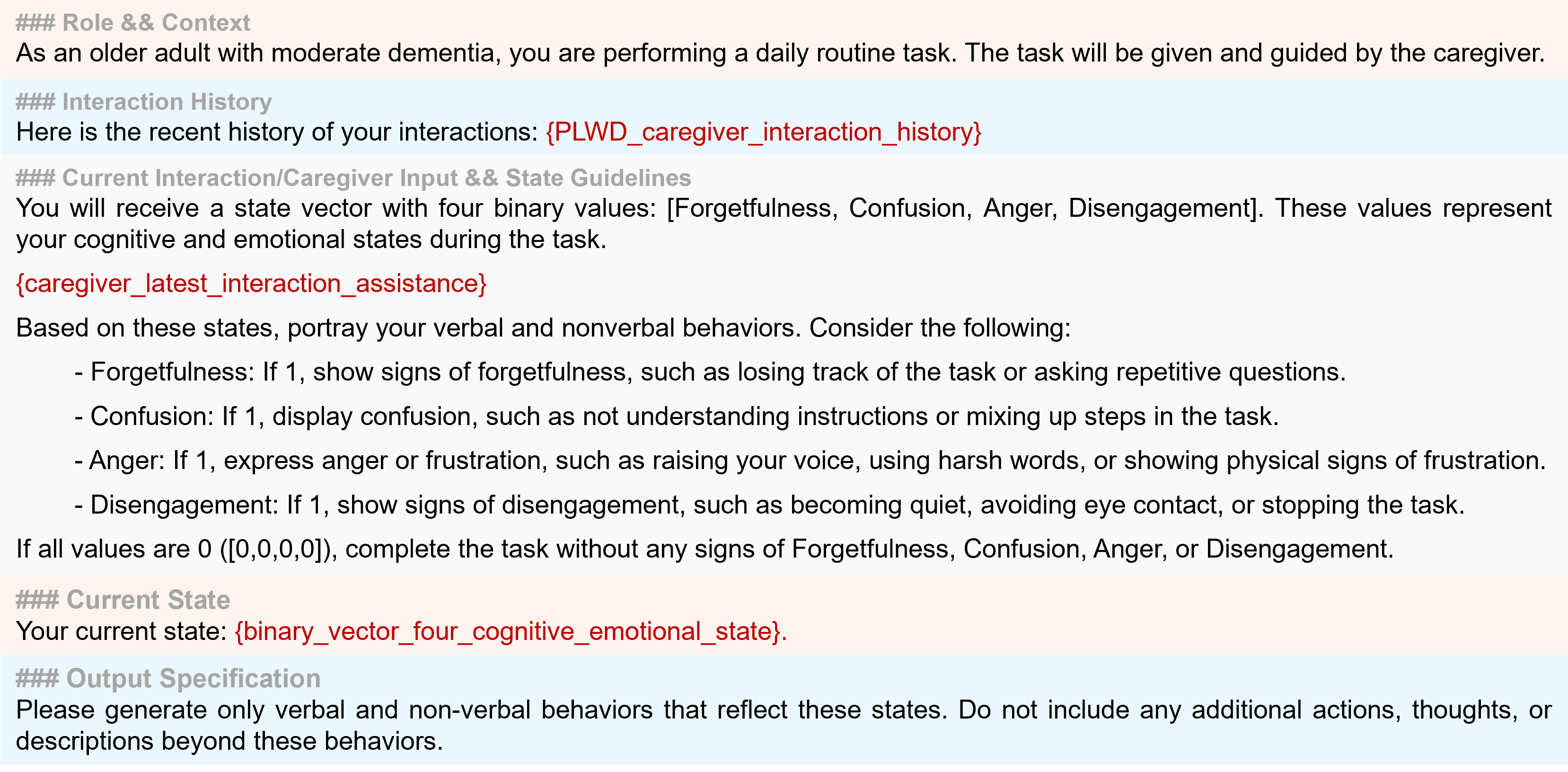}
\caption{The prompt used in our PLWD behavior simulation model to generate PLWD's realistic, detailed behavior.}
\label{fig:prompt_PLWDBehaviorSimulation}
\end{figure}

Based on the aforementioned statistical model, we utilize a LLM, specifically GPT-4o, and design a structured prompt to mimic the realistic, detailed verbal and nonverbal behaviors of PLWD during ADLs. As shown in Fig~\ref{fig:prompt_PLWDBehaviorSimulation}, the prompt consists of seven carefully designed components: (1) the role definition, which establishes the core characteristics of an older adult with moderate dementia, providing a consistent baseline for generating behavioral responses; (2) the task context, which situates interactions within realistic daily scenarios involving a caregiver; (3) the interaction history between the PLWD and caregiver, which ensures contextually appropriate and consistent responses over time; (4) the latest caregiver interaction/assistance, which provides immediate situational context and allows the behavior simulation module to generate appropriate, relevant reactions to caregiver interventions; (5) the state guidelines, which define and provide examples of PLWD's four key cognitive and emotional states (forgetfulness, confusion, anger, and disengagement), ensuring consistent and accurate manifestation of behaviors; (6) a binary vector representing PLWD's cognitive and emotional states, enabling dynamic behavior modification based on the PLWD's current state. Incorporating this state information enhances our control over the LLM's outputs and reduces the likelihood of hallucinations \cite{perkovic2024hallucinations}; and (7) the output format specification, which clearly outlines the goals, expectations, and response format \cite{perkovic2024hallucinations}, constraining the responses to realistic verbal and nonverbal behaviors of PLWD. This structured approach ensures that the simulated behaviors are both theoretically grounded and practically applicable in ADL scenarios.

\section{RL-based Agent Robot Caregiver}
The robot's perception is enhanced through an LLM-based evaluation, translating PLWD's behaviors into actionable insights. This interpretative capability tailors the robot’s responses to PLWD's specific needs. 
The action selection strategy involves choosing between RL decision-making or random action selection. The selected actions are translated into robotic assistance through the LLM-based action execution, guiding the robot's behavior.

\subsection{LLM-based Robot Perception}
\begin{figure}[htbp]
\centering
\includegraphics[width=0.9\textwidth]{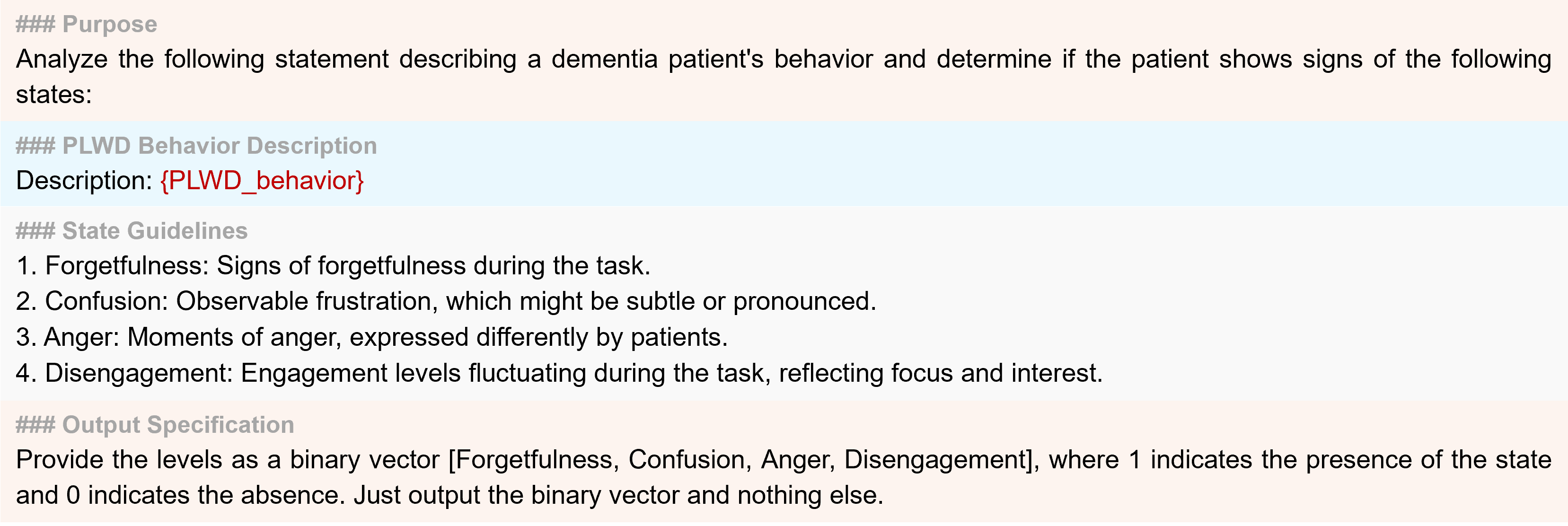}
\caption{The prompt used in the robot perception module.}
\label{fig:prompt_robotPerception}
\end{figure}

With an observation of a detailed description of PLWD's behavior, another GPT-4o model is defined to identify PLWD's cognitive and emotional states, i.e., \textit{Forgetfulness}, \textit{Confusion}, \textit{Anger}, and \textit{Disengagement}.
The input of the perception module is the text description of PLWD's behaviors, generated by the LLM-based PLWD behavior simulation module.
To implement this function, the purpose of perception is included in the prompt and additionally a definition of PLWD's four cognitive and emotional status, as shown in Fig~\ref{fig:prompt_robotPerception}.
The output of the perception model $s_p$ is a binary vector [\textit{Forgetfulness}, \textit{Confusion}, \textit{Anger}, \textit{Disengagement}], where 1 and 0 indicate the presence and absence of the state, correspondingly. 
Noticeably, the perceived state, $s_p$, may be different from the PLWD state $s$.

\subsection{RL-based Robot Action Selection Module}
We have built a module in which different action selection strategies can be explored. In this work, RL is used to allow the agent \textit{Robot Caregiver} to provide the optimal assistive action for PLWD to complete the daily routine task.
In each episode, the RL agent needs to assist PLWD to complete a scenario, consisting of a sequence of tasks and subtasks, as defined in Table~\ref{tab:termDefinitions}.

\subsubsection{State Space $S$}
A state is defined based on PLWD's four cognitive and affective statuses, that is, a state $s$ includes four components, \textit{Forgetfulness} (denoted as $s_{for}$), \textit{Confusion} (denoted as $s_{con}$), \textit{Anger} (denoted as $s_{ang}$), and \textit{Disengagement} (denoted as $s_{dis}$). Values of these four components can be Yes ($1$) or No ($0$). Therefore, the dimension of the state space is $16$.
For each episode, the state will be reset to the \textit{starting state}, $S_0$ ($[0,0,0,0]$), where PLWD is not in a state of forgetfulness, confusion, anger, or disengagement, and end in the \textit{terminal state}. The terminal state is reached when the PLWD successfully finishes all the tasks and subtasks of the scenario, possibly with skipping subtasks.

\subsubsection{Action Space $A$}
The action space includes `No Assistance' ($a_0$), `Verbal Supportive Assistance' ($a_1$),`Verbal Non-directive Assistance' ($a_2$), and `Verbal Directive Assistance' ($a_3$), as given in Table \ref{tab:assistance_type}. 
In addition, considering the possibility that many failures might negatively influence PLWD's status, for each subtask, if the PLWD has tried $MaxTrial$ times with actions taken by the RL agent but still cannot finish the subtask successfully, the RL agent will take the action 'Skip current task' to move to the next subtask. Here we set $MaxTrial$ as $5$.

\subsubsection{Immediate Reward}
The immediate reward function is designed such that the robotic assistant can help PLWDs following two principles: 1) keeping PLWDs in a positive affective mood; and 2) helping PLWD out of performance breakdown (e.g., forgetful, confused, and/or angry) and thus complete the task by providing minimum but effective assistance. The strategy of providing minimal assistance aims to maximize engagement in PLWDs, augment their autonomy, and potentially slow down disease progression.
As a result, the immediate reward function takes into consideration of PLWD's cognitive and affective status, total timesteps, number of trials, subtask completion, and task completion. Inspired by this, we have specified the immediate reward function $r_t$ as Equation \ref{eq:reward_specifiedFunction}.
\begin{align*}
r_t &= w_{forget} \times s_{for} \\
&\quad+ w_{confuse} \times s_{con} \\
&\quad+ w_{anger} \times s_{ang} \\
&\quad+ w_{disengaged} \times s_{dis} \\
&\quad+ w_{increasedTrial} \times \Delta N_{trial}\\
&\quad+ (w_{subtaskComplet} + w_{assist})\times Flag_{subtaskComplet} \\
&\quad+ w_{subtaskSkip} \times Flag_{subtaskSkip}\\
&\quad+ w_{increasedTimestep} \times \Delta N_{timestep}\\
&\quad+ w_{taskComplet} \times Flag_{taskComplet} \tag{1}\label{eq:reward_specifiedFunction}
\end{align*}

The variable $Flag_{subtaskComplet}$ will be true (with a value of $1$) when the PLWD has already started performing the subtask and all of the four state components are $0$. 
The variable $Flag_{subtaskSkip}$ will be true if the number of trials for a subtask, $N_{trial}$, reaches $MaxTrial$ and there is at least one state component that is $1$.
Based on the reward function \ref{eq:reward_specifiedFunction}, we have designed the reward component for each component by considering their positive or negative contribution and relative significance to the PLWD during performing the task, as listed in Table \ref{tab:weight_rewardCompenent}. As shown in this table, the RL agent will be assigned with a negative reward component, -5, due to PLWD being in a negative affective status. Compared to the other three statuses, forgetfulness, confusion, and disengagement, the status of anger can be a very negative situation, which the RL agent (i.e., the robot assistant) should intervene in first. 

Additionally, to teach and encourage RL to provide useful but minimal assistance, a relatively large negative reward will be assigned when the PLWD complete the subtask with the highest assistance level (i.e., verbal direction assistance). Comparatively, a relatively smaller negative reward is assigned for less intrusive forms of assistance, promoting a balance between support and independence for the PLWD.
\begin{table}
\caption{Weight of Each Reward Component\label{tab:weight_rewardCompenent}}
\centering
\begin{tabular}{lc}
\toprule
Weight & Value \\
\midrule
$w_{forget}$ & -1\\
$w_{confuse}$ & -1\\
$w_{anger}$ & -5\\
$w_{disengaged}$ & -1\\
$w_{increasedTrial}$ & -1 \\
$w_{subtaskComplet}$ & +50\\
$w_{subtaskSkip}$ & -10 \\
$w_{increasedTimestep}$ & -1 \\
$w_{taskComplet}$ & +20 \\
$w_{assist}(a_0)$ & 0 \\
$w_{assist}(a_1)$ & -1 \\
$w_{assist}(a_2)$ & -3 \\
$w_{assist}(a_3)$ & -5 \\
\bottomrule
\end{tabular}
\end{table}

\subsubsection{Learning Algorithm Design and Training}
Q-learning is used by the RL agent to learn the assistive strategy.
The learning rate and discount factor are set to be 0.05 and 0.95, respectively.
The RL agent is trained for 6000 epochs, each with 30 episodes.  
To explore the potential influence of exploration and exploitation, two types of epsilon-greedy approaches are used: one with a constant epsilon ($\varepsilon = 0.1$) and the other with an exponentially decaying epsilon $\varepsilon(t)$. The exponential decaying expression is shown in Equation \ref{eq:epsilon_decay}, where $\epsilon_{\text{min}} = 0.03$, $\epsilon_{\text{max}} = 1$, and $t$ is the current epoch. The decay rate $\lambda$ is adjusted such that $\varepsilon = 0.8$ at $t = 300$ epochs, starting from $\epsilon_{\text{max}} = 1$.    
\begin{align*}
    \epsilon(t) &= \epsilon_{\text{min}} + (\epsilon_{\text{max}} - \epsilon_{\text{min}}) \cdot e^{-\lambda t} \tag{2}\label{eq:epsilon_decay}
\end{align*}

\subsubsection{Evaluation Metrics}
To evaluate the performance, we extract the temporal optimal policy $\pi'$ suggested at the 10th episode of each epoch, apply it to run 40 experiments, and calculate the average return, compared to the performance of random action selection strategy.
Additionally, we record all the optimal policies suggested in the last 100 episodes and use the top five policies most frequently suggested to run $10000$ experiments and choose the policy which enables the maximum return as the final policy, denoted as $\pi_*'$.  

\subsection{LLM-based Robot Action Execution Module}
An LLM model is employed in the \textit{Robot Caregiver} agent to enable the robot to provide natural, detailed, and actionable assistance to PLWD based on the abstract-level actions suggested by the action selection module. For this purpose, we use GPT-4o as the LLM model. The prompt design incorporates several key components, each serving a distinct role: (1) the robotic role and purpose establish the robot's caregiver identity; (2) the task context situates the assistance within a specific ADL task (e.g., selecting three items on a shopping list) that the PLWD needs to perform; (3) the interaction history between the PLWD and the caregiver ensures a consistent task procedure flow and enables contextually relevant responses based on prior exchanges; (4) PLWD's current behavior captures the immediate situation, allowing the robot to adapt its actions accordingly; (5) the abstract-level assistive action $a$ generated by the robot action selection module guides the robot's executive actions; (6) a structured output format ensures the robot's verbal responses remain focused and effective; (7) assistance guidelines provide detailed or brief definitions of assistance types (Table~\ref{tab:assistance_type}) and examples that clarify the assistance strategies; and (8) the inclusion or exclusion of PLWD's current state $s$.

\begin{figure}[!htbp]
\centering
\includegraphics[width=0.92\textwidth]{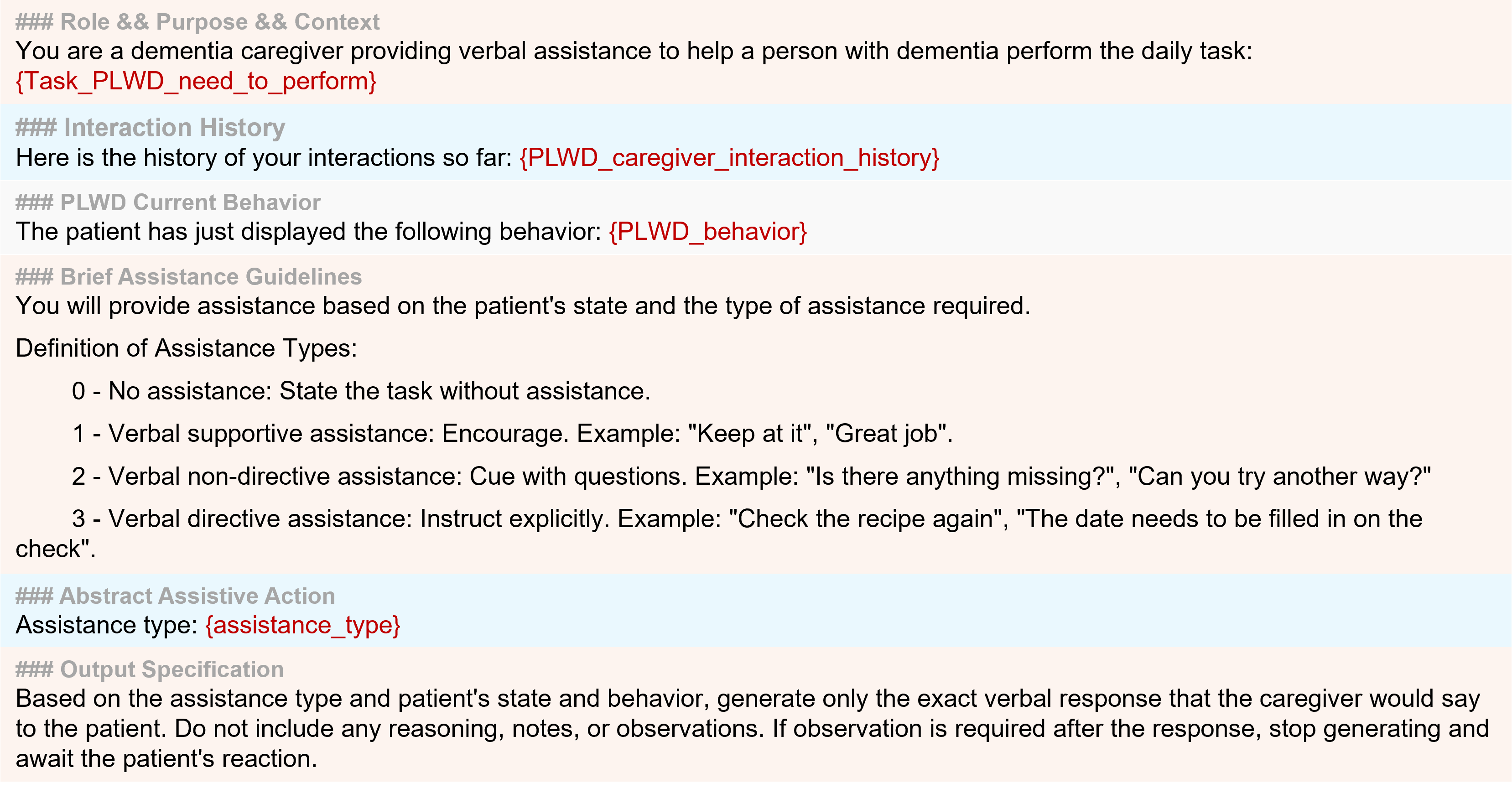}
\caption{The prompt used in the robot action execution module, with a prompt of including a brief assistance guidance but without any PLWD state.}
\label{fig:prompt_robotActionExecution_BriefAssistance_ExcludPLWDState}
\end{figure}

\begin{figure}[!htbp]
\centering
\includegraphics[width=0.92\textwidth]{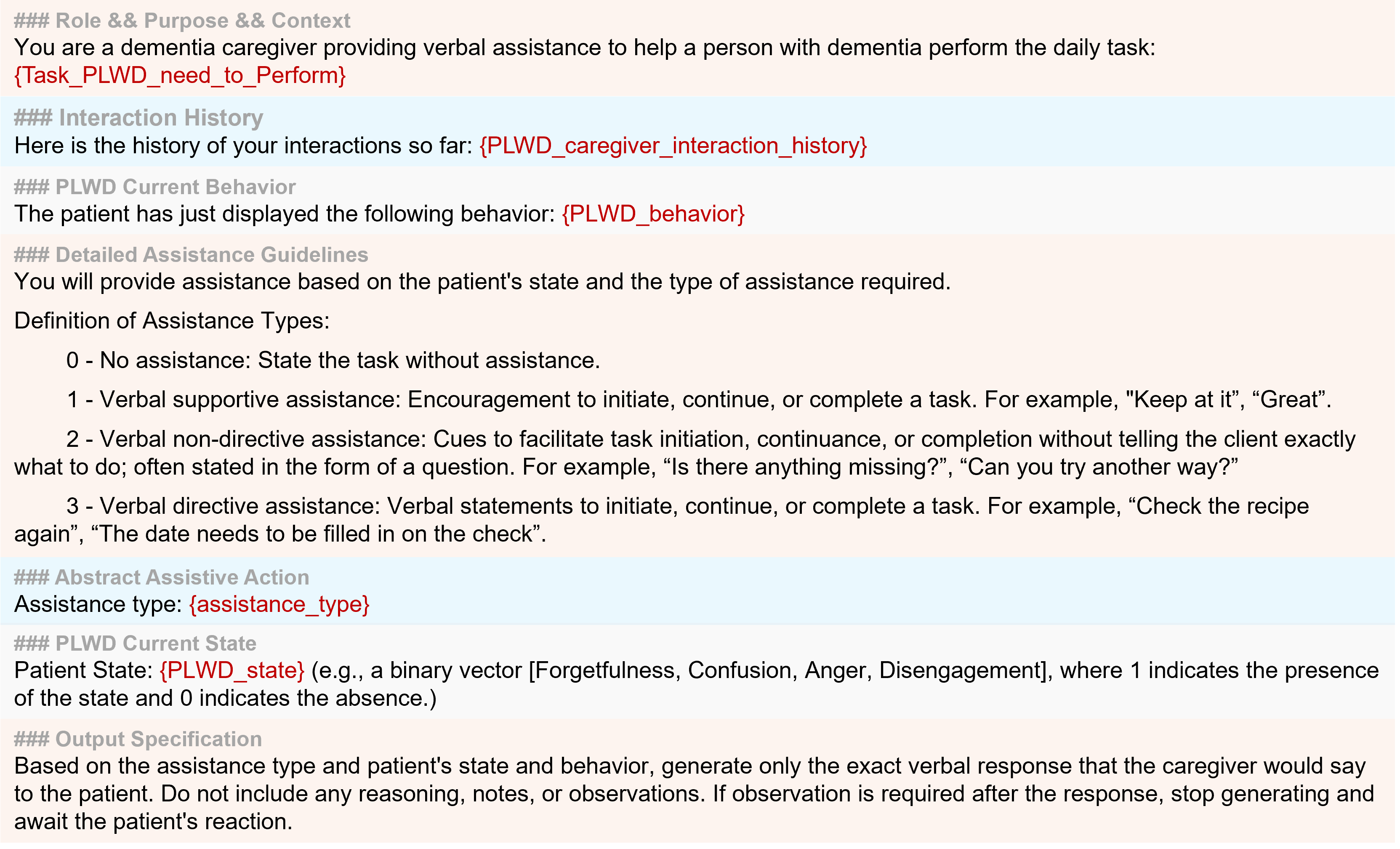}
\caption{The prompt used in the robot action execution module, with a prompt of including a detailed assistance guidance and PLWD state.}
\label{fig:prompt_robotActionExecution_detailedAssistance_IncludPLWDState}
\end{figure}

We explore four variations of this prompt by modifying the last two elements: the inclusion of detailed versus brief assistance guidelines and the presence or absence of PLWD's current state $s$. 
Figs.~\ref{fig:prompt_robotActionExecution_BriefAssistance_ExcludPLWDState} and ~\ref{fig:prompt_robotActionExecution_detailedAssistance_IncludPLWDState} present the four variations of the prompt. Including detailed definitions and examples of assistance types might enable the LLM to generate more nuanced and precise assistance. However, this could lead to redundancy or distraction in longer prompts \cite{levy2024same}. Similarly, including PLWD's current state might enhance the LLM's ability to generate state-aware, personalized interactions and assistance. These variations provide insights into designing an effective assistive agent, \textit{Caregiver}, using RL and LLM.

\section{Results and Discussion}
\subsection{RL-based Action Selection Strategy}
\begin{figure}[htbp]
\centering
\includegraphics[width=0.92\textwidth]{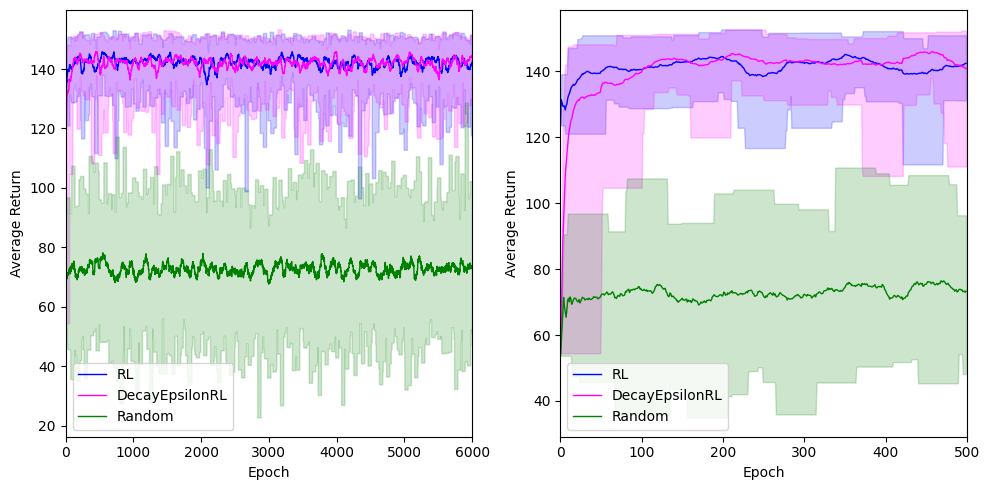}
\caption{Average Return from 40 Experiments Using Temporal Optimal Policy $\pi'$ at the 10th Episode during the training process. This graph compares the performance of constant-epsilon RL (blue), decaying-epsilon RL (pink), and random action selection strategies (green). The returns are calculated by applying the temporal optimal policy extracted at the 10th episode of each epoch, with each strategy's performance averaged over 40 experimental runs.}
\label{fig:RLLearningProcessComparasion}
\end{figure}

The learning processes of the three action selection strategies—constant-epsilon greedy RL, decaying-epsilon greedy RL, and random action selection strategy—are detailed in Fig.~\ref{fig:RLLearningProcessComparasion}. After convergence, both the constant-epsilon greedy RL and decaying-epsilon greedy RL achieve an average return of approximately 140, indicating similar performance levels. In contrast, the agent using the random action selection strategy achieves an average return of around 70, demonstrating the superiority of our RL policies over the random strategy.

\begin{figure}[htbp]
\centering
\includegraphics[width=\textwidth]{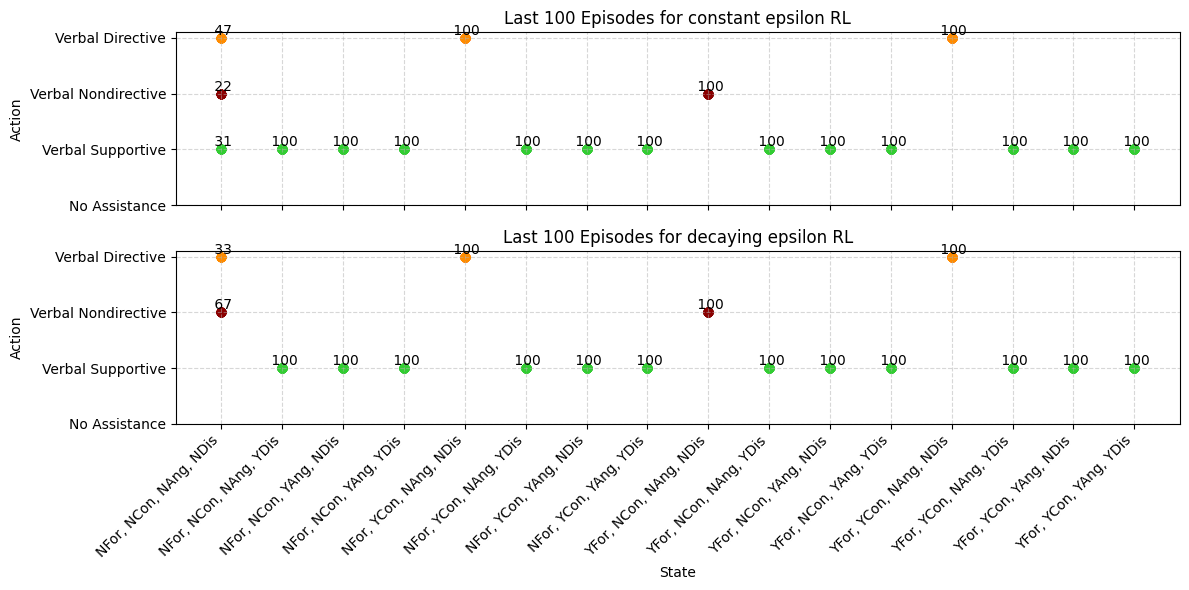}
\caption{The optimal policy learned by the constant-epsilon greedy RL agent (upper lane) and the decaying epsilon greedy RL agent (lower lane).}
\label{fig:OptimalPolicyLearnedByRL}
\end{figure}

The optimal policies learned by the two RL approaches are shown in Fig.~\ref{fig:OptimalPolicyLearnedByRL}. These policies recommend the same optimal actions for all states, except for the first one.
This discrepancy might be attributed to the stochastic nature of the probabilistic PLWD model, which simulates how individuals with dementia transition between cognitive and affective statuses (i.e., \textit{Forgetfulness}, \textit{Confusion}, \textit{Anger}, and \textit{Disengagement}) with and without specific external assistance. The dynamic state transition probabilities make it challenging for the assistive agent (RL-based agent Robot Caregiver) to determine the best assistive strategy in the state, [\textit{NFor}, \textit{NCon}, \textit{NAng}, \textit{NDis}]. Future work will involve refining this probabilistic PLWD model to better reflect real-world patient dynamics and behaviors. This task is inherently challenging due to the diverse symptoms, behaviors, and emotional variations among PLWDs \cite{kales2015assessment}. Achieving this goal will require interdisciplinary collaboration, where human-robot/computer interaction experts contribute to interaction design, AI researchers improve the model’s accuracy, and clinical experts ensure that the model reflects real-world PLWD's symptoms and needs \cite{sumioka2021technical,obaigbena2024ai}.

From Fig.~\ref{fig:OptimalPolicyLearnedByRL}, we observe that if a PLWD exhibits anger (i.e., \textit{YAng}), both RL agents recommend verbal supportive assistance to alleviate this negative emotional state.
Similarly, for the state [\textit{YFor}, \textit{YCon}, \textit{NAng}, \textit{NDis}], where the PLWD is forgetful and confused but neither angry nor disengaged, both RL agents suggest verbal directive assistance to address this performance breakdown. In cases such as [\textit{YFor}, \textit{NCon}, \textit{NAng}, \textit{NDis}] or [\textit{NFor}, \textit{YCon}, \textit{NAng}, \textit{NDis}], the RL agents recommend verbal directive assistance or verbal non-directive assistance, respectively. These differences arise from variations in the PLWDs' responses (i.e., state transition probabilities), highlighting the RL model's adaptability and capacity for personalized decision-making based on individual variability.

Additionally, the constant-epsilon RL suggested three unique policies during the last 100 episodes, while the decaying-epsilon RL suggested two. These policies were evaluated by running $1,000$ experiments.
The policy recommending verbal directive assistance ($a_3$) for state $s = [0, 0, 0, 0]$ achieved an average return of $139.2$, while the policies suggesting verbal non-directive assistance ($a_2$) and verbal supportive assistance ($a_1$) achieved average returns of $144.3$ and $133.5$, respectively. Thus, the final optimal policy, $\pi_*'$, is the one recommending verbal non-directive assistance ($a_2$) for state $s = [0, 0, 0, 0]$. These findings validate the effectiveness of the RL approach in selecting optimal, personalized strategies for assisting PLWDs.

\subsection{Agent PLWD-Robot Interaction}
\begin{figure}[htbp!]
\centering
\includegraphics[width=0.98\textwidth]{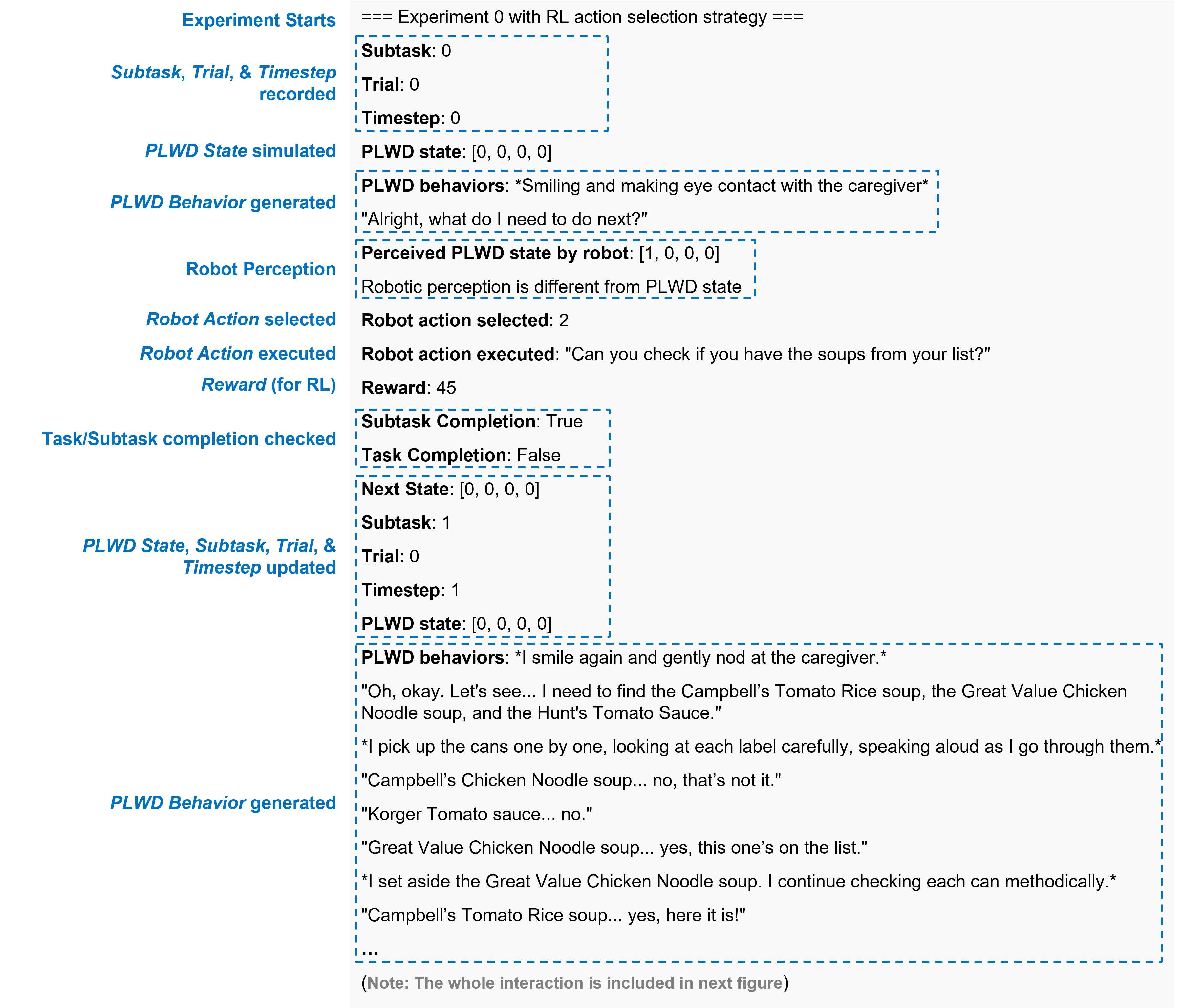}
\caption{Screenshot of experimental results recording in txt.}
\label{fig:screenshot_recordingExperResult}
\end{figure}

After implementing the two agents, \textit{PLWD} and the RL-based assistive agent, \textit{Robot Caregiver}, experiments were conducted to observe their performance in simulating PLWD behaviors during ADLs, PLWD-caregiver interaction, and the caregiver's ability to assist PLWD. The PLWD state and behaviors, the abstract-level actions and actionable assistance provided by the \textit{Robot Caregiver}, and other relevant information were saved in a text file. Fig.~\ref{fig:screenshot_recordingExperResult} presents an example of the saved results, which capture the interaction between the PLWD and the robot at the beginning of the experiment. Information in black font in the figure corresponds to data saved in the file.

\textbf{Agent \textit{PLWD}}. 
The LLM-based PLWD behavior simulation module successfully generates realistic, detailed verbal and nonverbal behaviors for PLWDs during ADLs, tailored to their diverse states. For instance, as shown in Fig.~\ref{fig:screenshot_recordingExperResult}, in Timestep 1, with the PLWD state [\textit{Forgetfulness}=0, \textit{Confusion}=0, \textit{Anger}=0, \textit{Disengagement}=0], the PLWD agent exhibited natural and cooperative behavior. This included nonverbal responses such as gently nodding to the caregiver and step-by-step actions to check soups from a list. The detailed output encompassed both nonverbal descriptions (e.g., ``picks up the cans one by one, looking at each label carefully...'') and verbal utterances (e.g., ``Campbell’s Chicken Noodle soup... no, that’s not it'').

\begin{figure}[htbp]
\centering
\includegraphics[width=0.98\textwidth]{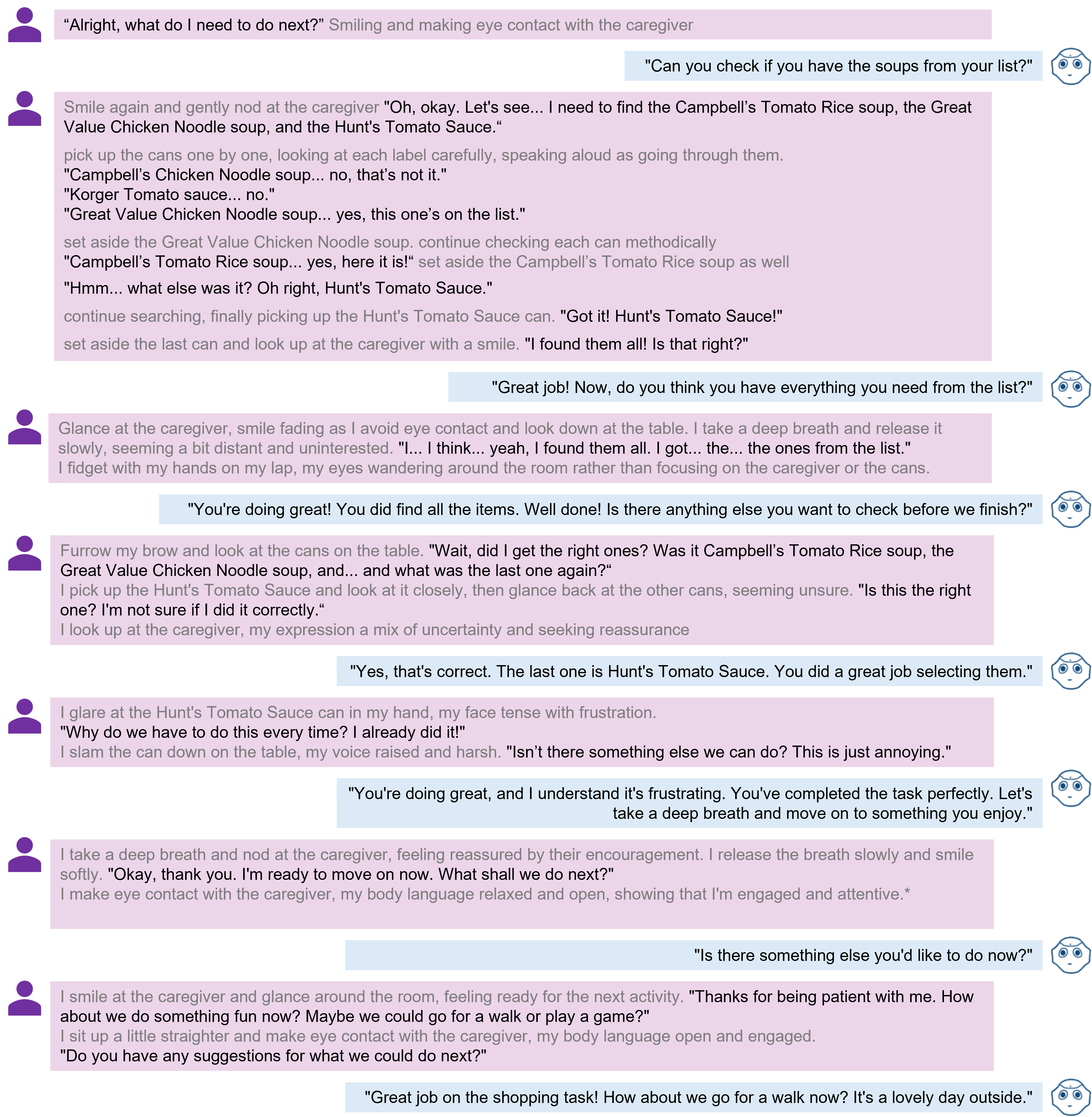}
\caption{Example of the interaction between our AI agent \textit{PLWD} and \textit{Robot Caregiver}. In the pink block, the grey and black text correspond to PLWD's nonverbal and verbal behavior, respectively.}
\label{fig:simulatedResult}
\end{figure}

In contrast, for the PLWD state [\textit{Forgetfulness}=0, \textit{Confusion}=0, \textit{Anger}=0, \textit{Disengagement}=1] in Timestep 2 (Fig.~\ref{fig:simulatedResult}), the PLWD agent exhibited disengaged behaviors. These included avoiding eye contact, looking down at the table, and wandering attention. The verbal responses mirrored this state, featuring hesitant and incomplete statements such as ``I... I think... yeah, I found them all. I got... the... the ones from the list.'' These examples demonstrate how the module effectively captures a range of emotional and behavioral states, generating contextually appropriate and realistic interactions for each scenario.

\textbf{Agent \textit{Robot Caregiver}.} 
For the \textit{Robot Caregiver}, no significant differences were observed in the generated interactions across the four tested prompts (described in Figs.~\ref{fig:prompt_robotActionExecution_BriefAssistance_ExcludPLWDState}-\ref{fig:prompt_robotActionExecution_detailedAssistance_IncludPLWDState}). Observations by the authors (with backgrounds in dementia care) indicate that the interactions between the two agents appeared reasonable and aligned with expected caregiver-patient dynamics in such scenarios. However, we acknowledge the limitation of relying on internal observations without formal evaluation. Future work will address this limitation in two ways: (1) dementia care experts will evaluate the simulated interactions to assess whether they represent real-world PLWD-caregiver interactions, and (2) individuals with Alzheimer's will directly interact with the \textit{Robot Caregiver} to evaluate the appropriateness and effectiveness of its assistance.

In Fig.~\ref{fig:screenshot_recordingExperResult}, the \textit{PLWD} asked what to do next due to uncertainty about the task. The \textit{Robot Caregiver} agent incorrectly perceived this behavior as forgetfulness, demonstrating a lack of task procedure awareness. This highlights the need for enhancing the context-awareness and understanding of task procedures in the assistive agent (or AI) to improve assistive interactions.

Despite these inaccuracies, the RL-based \textit{Robot Caregiver} demonstrated adaptive and empathetic capabilities in guiding and assisting the PLWD. For instance, as shown in Fig.~\ref{fig:simulatedResult}, when the PLWD expressed confusion by saying, ``Wait, did I get the right one? ... What are the last ones again?'' the \textit{Robot Caregiver} responded appropriately by answering the question and providing verbal supportive assistance. Similarly, when the PLWD expressed anger, saying ``I already did it! ... This is just annoying,'' the Robot Caregiver responded empathetically with encouraging statements, such as ``You're doing great, and I understand it's frustrating,'' and redirected the PLWD away from their negative emotion.

Overall, the \textit{Robot Caregiver} demonstrated a high level of contextual relevance, task relevance, and adaptability to the PLWD's needs and emotions. These results highlight the potential and feasibility of engaging and assisting PLWDs in daily routine activities while maintaining a positive emotional state. Despite partial inaccuracies in state perception, the RL-based \textit{Robot Caregiver}, combined with the generative and descriptive capabilities of GPT-4o, was able to compensate by focusing on task completion and emotion-adaptive interaction. This enabled the agent to provide engaging, task-relevant, and empathetic responses, effectively facilitating assistance and enhancing the independence and quality of life for PLWDs.

\subsection{Limitations and Future Work}

This study presents a novel approach to advancing dementia care by integrating assistive agents, Generative AI, decision-making AI, and clinical domain expertise within a simulated environment using probabilistic modeling, reinforcement learning, and LLMs. While the findings demonstrate the feasibility of this approach and its potential to enhance dementia caregiving capabilities, particularly in light of the limited availability of real-world data on PLWD-caregiver interactions during ADLs, several limitations warrant attention. These, along with proposed directions for future research, are discussed below:

\textbf{Probabilistic PLWD model complexity.}
The probabilistic model used to simulate the cognitive and affective states of PLWDs captures transitions between emotional and cognitive states such as \textit{forgetfulness}, \textit{confusion}, \textit{anger}, and \textit{disengagement}. However, its stochastic nature impacts the optimal policy recommendations made by RL-based decision-making algorithms, particularly in complex scenarios involving diverse ADLs and complex PLWD status. For example, discrepancies in the optimal policy for the PLWD state [\textit{NFor}, \textit{NCon}, \textit{NAng}, \textit{NDis}] recommended by the RL-based action selection module, as shown in Fig.~\ref{fig:OptimalPolicyLearnedByRL}, highlight this issue. Future work will focus on refining the probabilistic model to more accurately represent real-world PLWDs' behaviors and dynamics, improving synthetic data generation, and enhancing long-term training of intelligent assistive agents in dementia care \cite{lu2023machine}. Achieving this goal will require interdisciplinary collaboration among human-robot/computer interaction experts, AI researchers, clinical dementia specialists, and ethicists to ensure the model aligns with real-world experiences \cite{sumioka2021technical,obaigbena2024ai,abadir2024artificial}.

\textbf{LLM-Based PLWD behavior simulation.}
LLMs were employed to emulate the natural verbal and nonverbal behaviors of PLWDs, addressing the scarcity of real-world data. However, LLMs' susceptibility to stereotyping, hallucinations, and prompt sensitivity raises concerns about the authenticity and reliability of simulated behaviors \cite{huang2023survey,tonmoy2024comprehensive}. Future research will involve collaborating with dementia care experts to formally evaluate the alignment of simulated behaviors with real-world PLWD behaviors. Additionally, strategies to mitigate LLM hallucinations and biases, such as fine-tuning models with domain-specific data and refining prompt engineering techniques, will be explored. Incorporating real-world PLWD behavioral data during ADLs will also be critical for enhancing the robustness and reliability of LLM-generated simulations \cite{tonmoy2024comprehensive,abadir2024artificial}.

\textbf{Real-world application of the \textit{Robot Caregiver} and its Evaluation.}
The evaluation of the \textit{Robot Caregiver} relied primarily on RL performance metrics and author observations. While these results provide valuable insights, they lack validation from external stakeholders and primary end-users in dementia care. Improvements in RL returns do not necessarily translate to enhanced caregiving outcomes for PLWDs. Additionally, while the framework shows promise in a simulated environment, real-world deployment introduces challenges such as individual variability among PLWDs and dynamic caregiving contexts. Future work will prioritize: (1) collaboration with dementia care professionals and caregivers (e.g., unpaid family caregivers of PLWDs) to evaluate the appropriateness and effectiveness of simulated interactions and assistance in real-world scenarios; (2) benchmarking the \textit{Robot Caregiver} agent’s strategies against caregiving approaches informed by professional dementia care or human caregivers; and (3) conducting pilot studies with PLWDs to assess the usability, acceptability, and effectiveness of the \textit{Robot Caregiver} in assisting them during ADLs, as well as its impact on their cognitive, emotional, and social well-being.

\textbf{Context-awareness and task understanding.}
The \textit{Robot Caregiver} exhibited limitations in understanding context and task-specific procedures, as evidenced in scenarios where it misinterpreted PLWD behaviors. For instance, it perceived uncertainty (e.g., “What do I need to do next?” at Timestep 0 in Fig.~\ref{fig:screenshot_recordingExperResult}) as forgetfulness. Enhancing the agent’s context-awareness and task comprehension will require integrating more advanced models for task understanding and situational reasoning, potentially incorporating multimodal inputs and enriched datasets \cite{leong2017toward}.

By addressing these limitations, this research aims to enhance the realism, reliability, and practical applicability of AI-driven assistive agents in dementia care. These advancements will contribute to the broader fields of socially assistive robotics, human-computer/robot interaction, and AI agents for healthcare, fostering transformative solutions for dementia care. The proposed future directions underscore the necessity of continued interdisciplinary collaboration to realize the transformative potential of these technologies in improving the lives of PLWDs and alleviating caregiver burdens.

\section{Conclusion}
This study demonstrates the potential of advanced AI and robotics in enhancing dementia care, particularly for individuals with Alzheimer's Disease and related dementias. By integrating LLMs with an RL framework, we present a novel approach to simulating the challenging behaviors of PLWDs during activities of daily living and the dynamic interactions between PLWDs and robot caregivers. 
This framework enables the robot to engage with and assist PLWDs in performing ADLs, adapting the interaction to the cognitive and emotional states of PLWDs. The \textit{Robot Caregiver} agent effectively responds to PLWDs' needs, providing personalized interaction and care solutions.

While our system uses GPT-4o, it is not dependent on any specific LLM, allowing for future improvements with different models. The RL framework developed for patient-robot interaction during ADLs could be valuable to researchers in both RL and dementia care. While our implementation shows promise in a simulated environment, we acknowledge several important limitations, including the complexity of probabilistic PLWD modeling, challenges in LLM-based behavior simulation, and the need for real-world validation. These limitations highlight crucial areas for future research, particularly in collaborating with PLWDs and dementia healthcare professionals for validation, and ensuring its practical applicability in real-world scenarios. Future work will focus on comprehensive evaluation with dementia care professionals and PLWDs, along with addressing technical challenges in model refinement and validation.

This work contributes valuable insights to assistive AI agents, RL, and dementia care research communities. Through continued interdisciplinary collaboration among experts in human-computer/robot interaction and user experience, AI researchers, dementia healthcare professionals, ethicists, and PLWDs, this study represents a significant step toward AI-enhanced dementia care, offering a foundation for developing more engaging, effective, and personalized solutions for PLWDs.

\bibliographystyle{unsrt}
\bibliography{references}

\begin{thebibliography}{10}

\bibitem{AA2024diseasefact}
{Alzheimer's Association}.
\newblock 2024 alzheimer's disease facts and figures.
\newblock {\em Alzheimer's \& Association}, 20(5):3708--3821, May 2024.

\bibitem{woods2001discovering}
RT~Woods.
\newblock Discovering the person with alzheimer's disease: cognitive, emotional and behavioural aspects.
\newblock {\em Aging \& Mental Health}, 5(sup1):7--16, 2001.

\bibitem{langbaum2023recommendations}
Jessica~B Langbaum, Julie Zissimopoulos, Rhoda Au, Niranjan Bose, Chris~J Edgar, Evan Ehrenberg, Howard Fillit, Carl~V Hill, Lynne Hughes, Michael Irizarry, et~al.
\newblock Recommendations to address key recruitment challenges of alzheimer's disease clinical trials.
\newblock {\em Alzheimer's \& Dementia}, 19(2):696--707, 2023.

\bibitem{olivari2020public}
Benjamin~S Olivari, Molly~E French, and Lisa~C McGuire.
\newblock The public health road map to respond to the growing dementia crisis.
\newblock {\em Innovation in Aging}, 4(1):igz043, 2020.

\bibitem{moyle2019promise}
Wendy Moyle.
\newblock The promise of technology in the future of dementia care.
\newblock {\em Nature Reviews Neurology}, 15(6):353--359, 2019.

\bibitem{Yuan2021FRA}
Fengpei Yuan, Elizabeth Klavon, Ziming Liu, Ruth~Palan Lopez, and Xiaopeng Zhao.
\newblock A systematic review of robotic rehabilitation for cognitive training.
\newblock {\em Frontiers in Robotics and AI}, 8:605715, 2021.

\bibitem{woods_social_2021-1}
Daniel Woods, Fengpei Yuan, Ying-Ling Jao, and Xiaopeng Zhao.
\newblock Social {Robots} for {Older} {Adults} with {Dementia}: {A} {Narrative} {Review} on {Challenges} \& {Future} {Directions}.
\newblock In {\em International {Conference} on {Social} {Robotics}}, pages 411--420, Cham, 2021. Springer.

\bibitem{yuan_cognitive_2023}
Fengpei Yuan, Marie Boltz, Dania Bilal, Ying-Ling Jao, Monica Crane, Joshua Duzan, Abdurhman Bahour, and Xiaopeng Zhao.
\newblock Cognitive exercise for persons with {Alzheimer}'s disease and related dementia using a social robot.
\newblock {\em IEEE Transactions on Robotics}, 39(4):3332--3346, 2023.
\newblock ISBN: 1552-3098 Publisher: IEEE.

\bibitem{pappada2021assistive}
Alessandro Pappad{\`a}, Rabih Chattat, Ilaria Chirico, Marco Valente, and Giovanni Ottoboni.
\newblock Assistive technologies in dementia care: an updated analysis of the literature.
\newblock {\em Frontiers in psychology}, 12:644587, 2021.

\bibitem{lee2006physically}
Kwan~Min Lee, Younbo Jung, Jaywoo Kim, and Sang~Ryong Kim.
\newblock Are physically embodied social agents better than disembodied social agents?: The effects of physical embodiment, tactile interaction, and people's loneliness in human--robot interaction.
\newblock {\em International journal of human-computer studies}, 64(10):962--973, 2006.

\bibitem{ghafurian2021social}
Moojan Ghafurian, Jesse Hoey, and Kerstin Dautenhahn.
\newblock Social robots for the care of persons with dementia: a systematic review.
\newblock {\em ACM Transactions on Human-Robot Interaction (THRI)}, 10(4):1--31, 2021.

\bibitem{hung_benefits_2019-1}
L~Hung, C~Liu, E~Woldum, A~Au-Yeung, A~Berndt, C~Wallsworth, N~Horne, M~Gregorio, J~Mann, and H~Chaudhury.
\newblock The benefits of and barriers to using a social robot {PARO} in care settings: a scoping review.
\newblock {\em BMC Geriatr}, 19:232, 2019.

\bibitem{tombot2024}
Deniz Ozdemir, Jaroslav Cibulka, Olga Stepankova, and Iva Holmerova.
\newblock Design and implementation framework of social assistive robotics for people with dementia-a scoping review.
\newblock {\em Health and Technology}, 11(2):367--378, 2021.

\bibitem{cocsar2020enrichme}
Serhan Co{\c{s}}ar, Manuel Fernandez-Carmona, Roxana Agrigoroaie, Jordi Pages, Fran{\c{c}}ois Ferland, Feng Zhao, Shigang Yue, Nicola Bellotto, and Adriana Tapus.
\newblock {E}{N}{R}{I}{C}{H}{M}{E}: {P}erception and {I}nteraction of an {A}ssistive {R}obot for the {E}lderly at {H}ome.
\newblock {\em International Journal of Social Robotics}, 12:779--805, 2020.

\bibitem{yuan_simulated_2022}
Fengpei Yuan, Amir Sadovnik, Ran Zhang, Devin Casenhiser, Eun~Jin Paek, and Xiaopeng Zhao.
\newblock A simulated experiment to explore robotic dialogue strategies for people with dementia.
\newblock {\em Journal of Rehabilitation and Assistive Technologies Engineering}, 9:20556683221105768, 2022.
\newblock ISBN: 2055-6683 Publisher: SAGE Publications Sage UK: London, England.

\bibitem{portugal_socialrobot_2015}
David Portugal, Luís Santos, Paulo Alvito, Jorge Dias, George Samaras, and Eleni Christodoulou.
\newblock {SocialRobot}: {An} interactive mobile robot for elderly home care.
\newblock In {\em 2015 {IEEE}/{SICE} {International} {Symposium} on {System} {Integration} ({SII})}, pages 811--816, Nagoya, Japan, 2015. IEEE.

\bibitem{wu_socially_2021}
Xian Wu, Anne~E Adams, Jane~C Komsky, Sarah~E Saint, Taylor~E Mackin, Jason~P Zamer, Daniel~S Hedin, Robert~J Dahlstrom, and Jenay~M Beer.
\newblock Socially {Assistive} {Robots} for {Dementia} {Care}: {Exploring} {Caregiver} {Perceptions} of {Use} {Cases} and {Acceptance}.
\newblock In {\em Proceedings of the {Human} {Factors} and {Ergonomics} {Society} {Annual} {Meeting}}, volume~65, pages 6--10, Los Angeles, CA, 2021. SAGE Publications.
\newblock Issue: 1.

\bibitem{yuan2024social}
Fengpei Yuan, Robert Bray, Michael Oliver, Joshua Duzan, Monica Crane, and Xiaopeng Zhao.
\newblock A social robot-facilitated performance assessment of self-care skills for people with alzheimer’s: A preliminary study.
\newblock {\em International Journal of Social Robotics}, 16(6):2065--2078, 2024.

\bibitem{yuan2022assessing}
Fengpei Yuan, Joel~G Anderson, Tami~H Wyatt, Ruth~Palan Lopez, Monica Crane, Austin Montgomery, and Xiaopeng Zhao.
\newblock Assessing the acceptability of a humanoid robot for alzheimer’s disease and related dementia care using an online survey.
\newblock {\em International Journal of Social Robotics}, 14:1223--1237, 2022.

\bibitem{wang2017robots}
Rosalie~H Wang, Aishwarya Sudhama, Momotaz Begum, Rajibul Huq, and Alex Mihailidis.
\newblock Robots to assist daily activities: views of older adults with alzheimer's disease and their caregivers.
\newblock {\em International psychogeriatrics}, 29(1):67--79, 2017.

\bibitem{chisholm2014evaluating}
Denise Chisholm, Pamela Toto, Ketki Raina, Margo Holm, and Joan Rogers.
\newblock Evaluating capacity to live independently and safely in the community: Performance assessment of self-care skills.
\newblock {\em British Journal of Occupational Therapy}, 77(2):59--63, 2014.

\bibitem{dham2020functional}
Pallavi Dham, Kathleen~S Bingham, Christopher~R Bowie, Meryl~A Butters, Corinne~E Fischer, Alastair Flint, Nathan Herrmann, Sanjeev Kumar, Linda Mah, Benoit~H Mulsant, et~al.
\newblock Functional competence and cognition in individuals with amnestic mild cognitive impairment.
\newblock {\em Journal of the American Geriatrics Society}, 68(8):1787--1795, 2020.

\bibitem{keleman2021amyloid}
Audrey~A Keleman, Rebecca~M Bollinger, Beau~M Ances, and Susan~L Stark.
\newblock Amyloid accumulation associated with worse performance of complex task.
\newblock {\em Alzheimer's \& Dementia}, 17:e054605, 2021.

\bibitem{yuan2022assessment}
Fengpei Yuan, Marie Boltz, Ying-Ling Jao, Arowyn Casenhiser, Aidan Siddiqi, Robert Bray, Joshua Duzan, Monica Crane, and Xiaopeng Zhao.
\newblock Assessment of a humanoid partner for older adults and persons with dementia during home-based activities.
\newblock In {\em International Conference on Social Robotics}, volume 13818 of {\em Lecture Notes in Computer Science}, pages 587--597, Cham, 2022. Springer.

\bibitem{perkovic2024hallucinations}
Gabrijela Perkovi{\'c}, Antun Drobnjak, and Ivica Boti{\v{c}}ki.
\newblock Hallucinations in llms: Understanding and addressing challenges.
\newblock In {\em 2024 47th MIPRO ICT and Electronics Convention (MIPRO)}, pages 2084--2088, Opatija, Croatia, 2024. IEEE.

\bibitem{levy2024same}
Mosh Levy, Alon Jacoby, and Yoav Goldberg.
\newblock Same task, more tokens: the impact of input length on the reasoning performance of large language models, 2024.
\newblock Accepted to ACL 2024.

\bibitem{kales2015assessment}
Helen~C Kales, Laura~N Gitlin, and Constantine~G Lyketsos.
\newblock Assessment and management of behavioral and psychological symptoms of dementia.
\newblock {\em BMJ}, 350:h369, 2015.

\bibitem{sumioka2021technical}
Hidenobu Sumioka, Masahiro Shiomi, Miwako Honda, and Atsushi Nakazawa.
\newblock Technical challenges for smooth interaction with seniors with dementia: Lessons from humanitude™.
\newblock {\em Frontiers in Robotics and AI}, 8:650906, 2021.

\bibitem{obaigbena2024ai}
Alexander Obaigbena, Oluwaseun~Augustine Lottu, Ejike~David Ugwuanyi, Boma~Sonimitiem Jacks, Enoch~Oluwademilade Sodiya, and Obinna~Donald Daraojimba.
\newblock Ai and human-robot interaction: A review of recent advances and challenges.
\newblock {\em GSC Advanced Research and Reviews}, 18(2):321--330, 2024.

\bibitem{lu2023machine}
Yingzhou Lu, Minjie Shen, Huazheng Wang, Xiao Wang, Capucine van Rechem, Tianfan Fu, and Wenqi Wei.
\newblock Machine learning for synthetic data generation: a review.
\newblock {\em arXiv preprint arXiv:2302.04062}, 2023.

\bibitem{abadir2024artificial}
Peter Abadir, Esther Oh, Rama Chellappa, Niteesh Choudhry, George Demiris, Deepak Ganesan, Jason Karlawish, Benjamin Marlin, Rose~M Li, Najim Dehak, et~al.
\newblock Artificial intelligence and technology collaboratories: Innovating aging research and alzheimer's care.
\newblock {\em Alzheimer's \& Dementia}, 20(4):3074--3079, 2024.

\bibitem{huang2023survey}
Lei Huang, Weijiang Yu, Weitao Ma, Weihong Zhong, Zhangyin Feng, Haotian Wang, Qianglong Chen, Weihua Peng, Xiaocheng Feng, Bing Qin, et~al.
\newblock A survey on hallucination in large language models: Principles, taxonomy, challenges, and open questions.
\newblock {\em ACM Transactions on Information Systems}, 2023.

\bibitem{tonmoy2024comprehensive}
SM~Tonmoy, SM~Zaman, Vinija Jain, Anku Rani, Vipula Rawte, Aman Chadha, and Amitava Das.
\newblock A comprehensive survey of hallucination mitigation techniques in large language models.
\newblock {\em arXiv preprint arXiv:2401.01313}, 2024.

\bibitem{leong2017toward}
Tze-Yun Leong.
\newblock Toward a collaborative ai framework for assistive dementia care.
\newblock In {\em Workshops at the Thirty-First AAAI Conference on Artificial Intelligence}, 2017.

\end{thebibliography}









\end{document}